# Learning Membership Functions in a Function-Based Object Recognition System

**Kevin Woods**                                                          WOODS@BIGPINE.CSEE.USF.EDU
*Computer Science & Engineering*
*University of South Florida*
*Tampa, FL 33620-5399*

**Diane Cook**                                                           COOK@CENTAURI.UTA.EDU
*Computer Science & Engineering*
*University of Texas at Arlington*
*Arlington, TX 76019*

**Lawrence Hall**                                                        HALL@WATERFALL.CSEE.USF.EDU
**Kevin Bowyer**                                                         KWB@BIGPINE.CSEE.USF.EDU
*Computer Science & Engineering*
*University of South Florida*
*Tampa, FL 33620-5399*

**Louise Stark**                                                         STARK@NAPA.ENG.UOP.EDU
*Electrical and Computer Engineering*
*University of the Pacific*
*Stockton, CA 95211*

## Abstract

Functionality-based recognition systems recognize objects at the category level by reasoning about how well the objects support the expected function. Such systems naturally associate a "measure of goodness" or "membership value" with a recognized object. This measure of goodness is the result of combining individual measures, or membership values, from potentially many primitive evaluations of different properties of the object's shape. A membership function is used to compute the membership value when evaluating a primitive of a particular physical property of an object. In previous versions of a recognition system known as GRUFF, the membership function for each of the primitive evaluations was hand-crafted by the system designer. In this paper, we provide a learning component for the GRUFF system, called OMLET, that automatically learns membership functions given a set of example objects labeled with their desired category measure. The learning algorithm is generally applicable to any problem in which low-level membership values are combined through an *and-or tree* structure to give a final overall membership value.

## 1. Introduction

In any computer vision (CV) application involving the recognition or the detection of "objects", descriptions of the types of objects to be recognized are required. Object descriptions can be explicitly supplied by a human "expert". Alternatively, machine learning techniques can be used to derive descriptions from example objects.

There are some advantages to learning object descriptions from examples rather than from direct specification by an expert. Specifically, it may be difficult for a person to provide a CV system with an accurate description of an object that is general enough to cover the possible variations in the visual appearance of different instances of the object. For example, no two tumors in medical images will look exactly the same. Similarly, it would be cumbersome for a human to provide a CV system with the ranges of possible values for all the different physical aspects of chairs (i.e., What are the possible surface areas of the seating surface of a chair? How is the seating surface supported?). Considerable "tweaking" of the object description parameters may be required by a human expert in order to achieve satisfactory system performance. Machine learning techniques can be used to generate concepts that are consistent with observed examples. Some examples of such learning systems include C4.5 (Quinlan, 1992), and AQ (Michalski, 1983). System performance is affected by the ratio of the number of training examples to the number of features used to describe the examples, and the accuracy with which the examples represent the "real-world" objects the CV system may encounter.

A function-based object recognition system is an example of a CV system for which machine learning techniques can be useful in the development of object descriptions. A function-based object recognition system recognizes an object by classifying it into one or more generic object categories which describe the function that the object might serve (Bogoni & Bajcsy, 1993; Brand, 1993; Di Manzo, Trucco, Giunchiglia, & Ricci, 1989; Kise, Hattori, Kitahashi, & Fukunaga, 1993; Rivlin, Rosenfeld, & Perlis, 1993; Stark & Bowyer, 1991, 1994; Sutton, Stark, & Bowyer, 1993; Vaina & Jaulent, 1991). Each object category is defined in terms of the functionality required of an object that belongs to the category. For example, an object category might be defined as:

$$straight\_back\_chair ::= provides\_sittable\_surface~\&~provides\_stability~\&~provides\_back\_support$$

indicating that an object can be classified as a straight back chair to the degree that it satisfies the conjunction of the three functional properties.

The functional properties are themselves defined in terms of primitive evaluations of different aspects of an object's shape. For example, candidate surfaces may be checked for *provides_sittable_surface* by evaluating whether they have appropriate width, depth and height above the support plane. In many cases, there is not a unique ideal value for some given aspect of an object's shape, but instead there is a range of values that can be considered equivalent in terms of "goodness". For example, anything between 0.45 to 0.55 meters might be an equally acceptable height for a seating surface. However, as a particular shape measurement becomes too small or too large, the evaluation measure should be reduced. Fuzzy set theory provides a mathematical framework for handling this "goodness of fit" concept. In our case, a fuzzy membership function transforms a physical measurement (i.e., height of an object's surface above the ground) into a membership value in the interval [0,1]. This membership value, or evaluation measure, denotes the degree to which the object (or portion of the object) fits the primitive physical concept (i.e., how well the height of the surface matches the seating surface height of typical chairs). Thus, a separate measure of goodness is produced for each primitive evaluation. These measures are combined to produce a final aggregate measure of goodness for the object.

The GRUFF system (Stark & Bowyer, 1991) is a function-based object recognition system which utilizes fuzzy logic, in the manner just described, to evaluate 3-D shapes. In previous versions of GRUFF, the fuzzy membership functions embedded in the system have been collectively hand-crafted and refined to produce the best results over a large set of example shapes. These membership functions are ideal candidates to be learned from examples using a machine learning approach.

In this paper, we present a method of automatically learning the collection of fuzzy membership functions from a set of labeled example shapes. Due to the system constraints imposed by GRUFF, general-purpose machine learning algorithms, such as neural networks, genetic algorithms, or decision trees, are not readily applicable. Thus, a new special-purpose learning component, called OMLET, has been developed. OMLET is tested with synthetic data for two different object categories (chairs and cups), and with data collected from human evaluations of physical chairs. Results are presented to show that (a) learning the membership functions in this way provides a level of recognition performance equivalent to that obtained from the "hand-tweaked" GRUFF, and (b) the learning method is compatible with human interpretation of the shapes. The approach should be generally applicable to any system in which a set of primitive evaluation measures is combined to produce an overall measure of goodness for the final result.

This paper is organized as follows. Section 2 discusses some related work, and justifies our need to develop a special-purpose learning component. Section 3 introduces the GRUFF object recognition system. Section 4 presents the new learning component, called OMLET. At this point, we should state that the material in Section 3 has previously been published, and is presented here to facilitate an understanding of the new learning component. Although OMLET has been specifically "tailored" as an add-on learning component for the GRUFF system, it applies to a data structure that can be used in other systems. In general, OMLET can be described as a system for learning in the context of a fuzzy AND/OR categorization tree. We point the reader with any questions concerning GRUFF'S object recognition paradigm to the references provided. Section 5 describes our experimental design and the data sets that are utilized. Section 6 documents the experimental results and gives our analysis of them. Finally, in Section 7 a summary of the paper is given and conclusions are drawn.

## 2. Related Work

There are two ways that learning might be used to ease the construction of systems such as GRUFF. The first is that the rules (or proof tree) that make up GRUFF could be built by an inductive learning system. C4.5, a decision tree learner (Quinlan, 1992), is a good example of this class of learning systems. However, these types of inductive classification systems cannot adequately replace the functionality of the GRUFF/OMLET system. OMLET allows examples which have less than perfect membership in a class to be used for training. There is no direct way to accomplish this in a system such as C4.5. A decision-tree based system would probably require different trees to be trained for parent and child categories. The functional concepts (*provides_sittable_surface*, for example) would get lost in the training process if the individual features for a chair were directly used. We could train a series of trees to learn functional concepts individually, then train a decision tree to combine the results. In such an approach the parameters of the membership functions that are learned in this paper would be learned implicitly in the construction of a decision tree for a functional concept and any resulting rules. Replacing GRUFF/OMLET with a decision tree or other general-purpose rule learner is possible, but would require extensive work to preserve the idea of functional object recognition.

OMLET is aimed at the second area in which a GRUFF-like system could benefit from learning, which is in tuning the membership functions. A knowledge primitive might be a sittable surface. Given measurements for a specific surface of an object in a specific orientation, it is necessary to develop a representation of acceptable bounds on the measurements to determine whether the surface has the area to be sittable.

Techniques from other areas of machine learning have been used to represent and learn probabilistic and fuzzy membership functions. For example, belief networks provide a mechanism for representing probabilistic relationships between features of a domain. Individual feature probabilities can be combined to generate the probability of a complex concept by propagating belief values and constraints through the network. Adaptive probabilistic networks are a kind of belief nets that can learn the individual probability values and distributions using gradient descent (Pearl, 1988; Cooper & Herskovits, 1992; Spiegelhalter, Dawid, Lauritzen, & Cowell, 1993). The structure of belief nets and their update algorithms are similar to the approaches found in OMLET. However, OMLET incorporates symbolic theorem proving, a feature that is fundamental to performing function-based object recognition, as well as value propagation.

Similar research has been performed to learn fuzzy membership functions using adaptive techniques such as genetic algorithms and classifier systems (Parido & Bonelli, 1993; Valenzuela-Rendon, 1991). Much of this work can only be used to learn individual membership functions and cannot handle combinations of input. Once again, little work has been directed at learning fuzzy memberships in the context of a rule-based system. Additional refinement techniques such as reinforcement learning (Mahadevan & Connell, 1991; Watkins, 1989), neural networks, and statistical learning techniques can also be used to refine confidence values.

This project represents a new direction in computer vision and machine learning research; namely, the integration of machine learning and computer vision methods to learn fuzzy membership functions for a function-based object recognition system. Although learning such functions in a rule-based context is a novel effort, similar research has been performed in the area of refining certainty factors for intelligent rule bases. For example, Mahoney and Mooney (1993) and Lacher et al. (1992) use backpropagation algorithms to adjust certainty factors of existing rules in order to improve classification of a given set of training examples. In contrast to OMLET's approach, all of these systems refine values that represent a measure of belief in a given result and are adjusted according to the combination functions of certainty factors. OMLET's measures represent degrees of fuzzy membership in an object class, and the refinement method propagates error through an AND/OR tree.

The work by Wilkins and Ma (1994) focuses on revising probabilistic rules in a classification expert system. Probabilistic weights are applied to each rule, indicating the strength of the evidence supplied by the rule. However, refinements to the rule occur in the form of modifying the applicability of the rule by generalizing, specializing, deleting or adding rules, instead of automatically refining the weight of the rule. The authors avoid automatic refinement of weights because the resulting rule base may not be interpretable by experts.

Towell and Shavlik (1993) convert a set of rules into a representation suitable for a neural net, then train the network and re-extract the refined rules. The initial network can be set up for a chain of rules. The extracted rules will not necessarily have the clear functional meaning that our approach aims at preserving.

There are several new approaches to learning and tuning fuzzy rules (Ishibuchi, Nozaki, & Yamamoto, 1993; Berenji & Khedkar, 1992; Jang, 1993; Jang & Sun, 1995) that use genetic algorithms or specialized kinds of neural networks, some making use of reinforcement learning. These approaches might provide an alternative way to learn the membership values provided the initial functional rules are given as fuzzy rules. However, some modifications to the learning approaches would be needed as they normally work in domains without rule chaining or hierarchies of rules as there are in Gruff/Omlet.

## 3. The Gruff Object Recognition System

The Gruff acronym stands for Generic Representation Using Form and Function (Stark & Bowyer, 1991). The Gruff recognition system takes a 3-D shape description as input, reasons about whether the shape could belong to any of the object categories known to Gruff, and outputs an interpretation for each category to which the object could belong. An "interpretation" is a specified orientation and a labeling of the parts of the shape which are identified as satisfying the functional properties. See Figure 1 for an example of an interpretation.

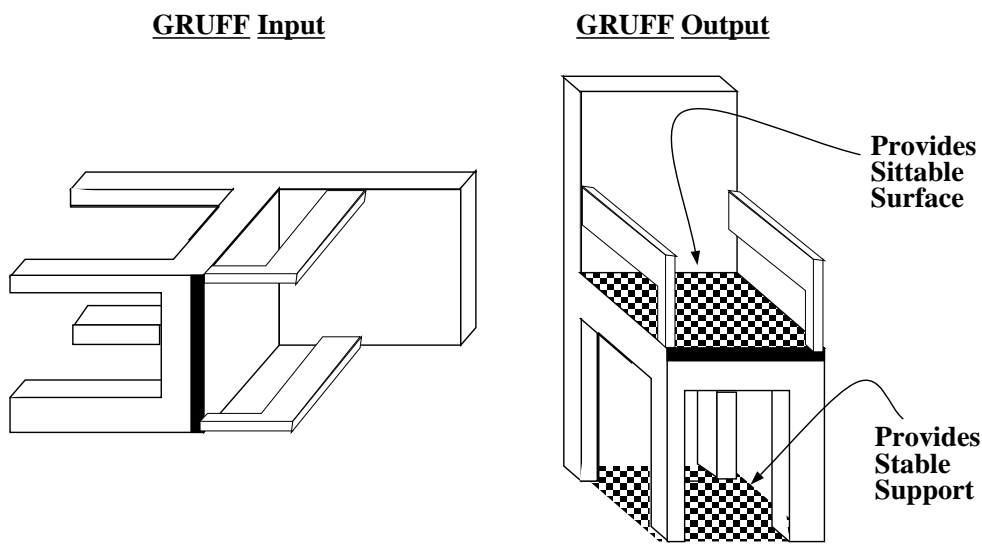

Figure 1: Gruff interpretation of a 3-D shape for the category *conventional chair*. Elements of the shape are labeled with the functional property they provide.

### 3.1 The Knowledge Primitives

All of GRUFF's reasoning about shape is performed using "low level" procedural knowledge which is implemented as a set of *knowledge primitives*. Each knowledge primitive represents some primitive physical property concerning shape, physics, or causation. Each knowledge primitive takes some (specified portions of a) 3-D shape description as its input, along with values of the parameters for the primitive, and returns an *evaluation measure* between 0 and 1. The evaluation measure represents how well the shape element satisfies the particular invocation of the primitive.

The knowledge primitives used by GRUFF to recognize chairs are (Stark & Bowyer, 1991, 1994; Sutton et al., 1993):

1. **relative_orientation (normal_one, normal_two, range_parameters)**

    This primitive determines if the angle between the normals for two surfaces (normal_one and normal_two) falls within a desired range.

2. **dimensions ( shape_element, dimension_type, range_parameters )**

    This primitive can be used to determine if the dimension (e.g. width or depth) of a surface lies within a specified range.

3. **proximity ( proximity_type, shape_element_one, shape_element_two )**

    This primitive can be used to check qualitative relations between shape_elements, such as *above*, *below* and *close to*.

4. **clearance ( object_description, clearance_volume )**

    This primitive can be used to check for a specified volume of unobstructed free space in a location relative to a particular part of the shape.

5. **stability ( shape, orientation, applied_force )**

    This primitive can be used to check that a given shape is stable when placed on a flat supporting plane in a given orientation and with a (possibly zero) force applied.

Each of the first two knowledge primitives include four range parameters: $z1$ (stands for 1st zero point), $n1$ (1st normal point), $n2$ (2nd normal point), and $z2$ (2nd zero point). These parameters are used to define a trapezoidal fuzzy membership function, as in Figure 2, for calculating an *evaluation measure* for the invocation of the primitive. The last three of the knowledge primitives do not have range parameters. They return an evaluation measure of 1 or 0 depending on whether or not the primitive physical property has been satisfied.

Trapezoidal membership functions reflect a desire to name (categorize) objects in a manner compatible with human naming. There is typically a non-trivial range for the "ideal" value of many physical properties related to functionality. For example, while there is a unique value for the mean sittable surface area of a population of chairs, that value is not the only one that would rate a perfect "1.0" for sittability. Reasonable deviations result in no decrease in the sittability. When the sittable surface area falls outside the ideal range (i.e., between $z1$ and $n1$, or between $n2$ and $z2$ in Figure 2), the evaluation measure is reduced, indicating the surface provides a less than perfect (but still functional) sittable area. Finally, when the area falls outside the range of values (less than $z1$, or greater than $z2$ in Figure 2), the surface can no longer function as the sittable portion of a chair, and a evaluation measure of 0 is returned.

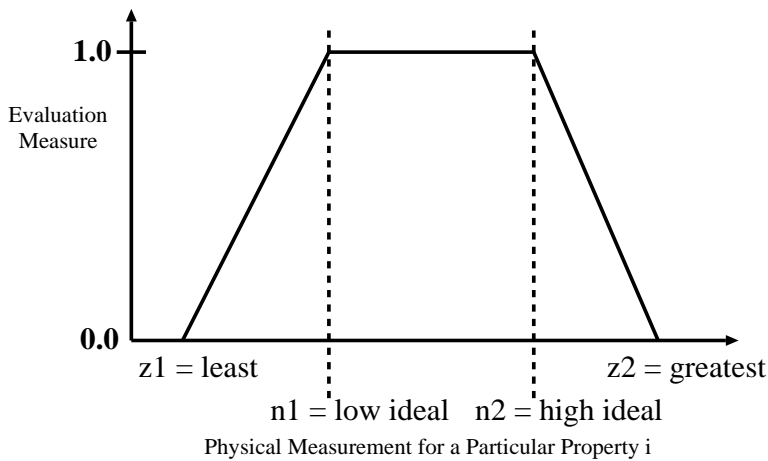

Figure 2: Fuzzy membership function returns an evaluation measure of a primitive physical property.

### 3.2 The Category Definition Tree

GRUFF's knowledge about different object categories is implemented as a *category definition tree*, the leaves of which represent invocations of the knowledge primitives. The category definition tree for the chair category is illustrated in Figure 3.

A node in a category definition tree may have two subtrees. One subtree gives the definition of the category in terms of a list of functional properties. In our chair example, an object must satisfy the functional properties of *stability* and *provides_sittable_surface* in order to be considered a member of the category *conventional chair*. Each functional property may be defined in terms of multiple primitives. The evaluation measures of individual primitives are combined (in a manner to be discussed shortly) to determine how well the functional properties have been satisfied. These functional property measures are further combined to arrive at an overall evaluation measure for a category node.

The other subtree defines a subcategory. A subcategory is a specialization of its parent (or superordinate) category, and thus provides a more detailed elaboration of the definition of its parent. A subcategory node has a subtree of functional properties that are required in addition to those of the parent category. For example, in Figure 3, the subcategory *straightback chair* is a specialization of a *conventional chair* with the additional functional requirement *provides_back_support*. The overall evaluation measure for a subcategory node is a combination of its parent category evaluation measure and the evaluation measure associated with the additional functional properties. In Figure 3, the overall measure for the subcategory *straightback chair* is a combination of the measures from the *conventional*

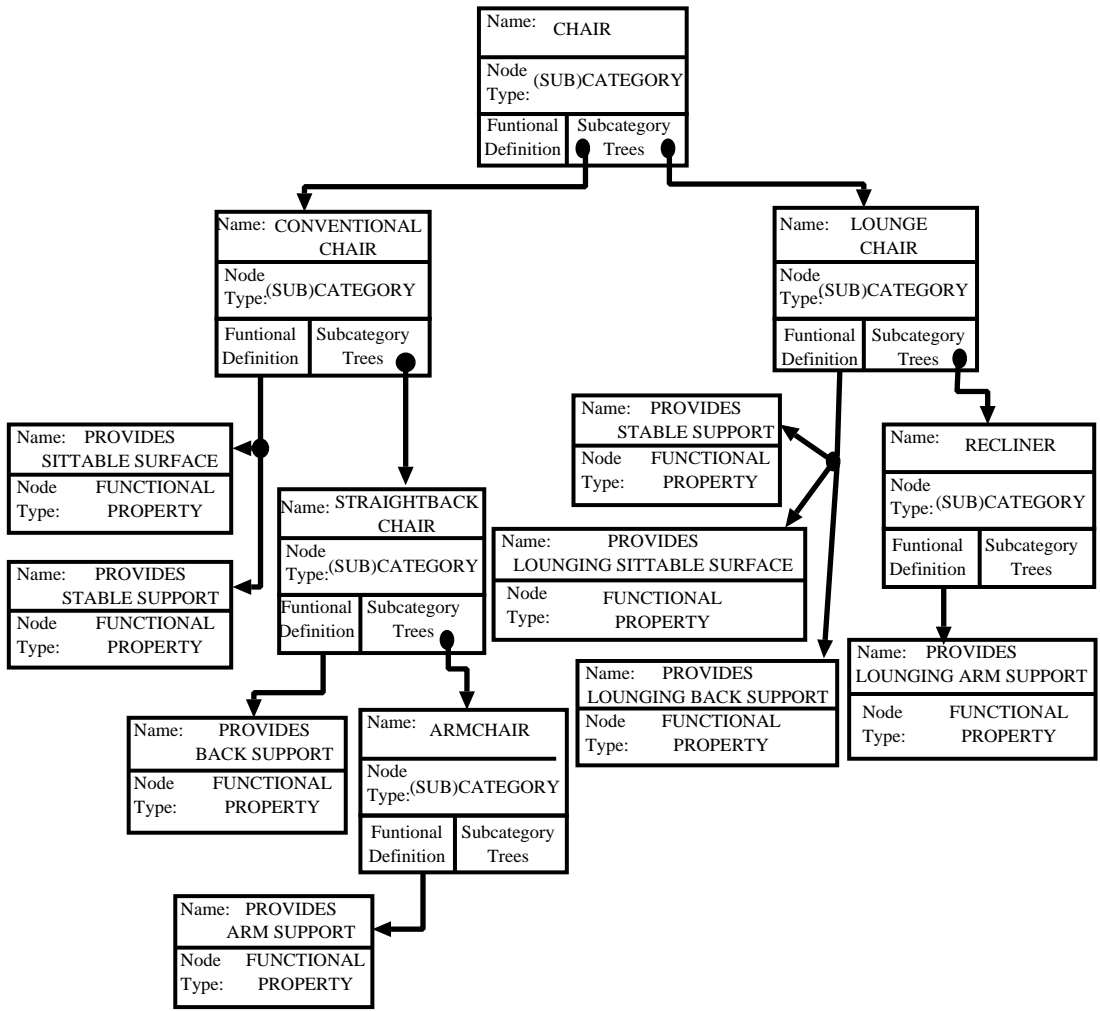

Figure 3: Category definition tree for the basic level category *chair*.

*chair* node and the *provides_back_support* subtree. Note that subcategory measurements do not contribute to the cumulative measure for a parent category. There may be multiple levels of subcategories, as with *conventional chair*, *straightback chair*, and *armchair* in Figure 3.

Category nodes which have no associated functional properties (such as the root node *chair* in Figure 3) do not have associated evaluation measures. These nodes are used to set up the control structure of the function-based definition. However, they do provide the category definition since an object that is a member of a subcategory is automatically a member of all its predecessor categories. For example, in Figure 3, an object that belongs to the subcategory *straightback chair* also belongs to the categories *conventional chair* and *chair*. A superordinate category *furniture* could be added above the *chair* category (Stark & Bowyer, 1994).

### 3.3 Combination of Evidence

The evaluation measures returned by the primitive invocations at a functional property node are combined using the T-norm:

$$T(a, b) = a \times b$$

where $a$ and $b$ are the measures being combined. This T-norm is commonly referred to as the *probabilistic and* (PAND) function (Bonissone & Decker, 1986). The immediate parent category node directly receives an associated measure by combining the measures of the functional property nodes using the same T-norm.

For example, the functional property *provides_sittable_surface* is defined by six primitives. For simplicity, we'll denote the evaluation measures returned by these six primitives as $p_1$ through $p_6$. The functional property *stability* is defined by a single primitive, which also returns an evaluation measure ($p_7$). To determine the overall evaluation measure of a shape for the category *conventional chair* we compute

$$conventional\ chair ::= provides\_sittable\_surface\ \text{PAND}\ stability$$

where

$$provides\_sittable\_surface ::= p_1\ \text{PAND}\ p_2\ \text{PAND}\ p_3\ \text{PAND}\ p_4\ \text{PAND}\ p_5\ \text{PAND}\ p_6$$

and

$$stability := p_7$$

Since the definition of a (sub)category is a conjunction of required functional properties, the cumulative measure should be dominated by the "weakest link" in the individual primitive evaluation measures, a property of the PAND function. So, an evaluation measure of 0 for any one primitive physical property will result in a cumulative evaluation measure of 0. An evaluation measure of 1 indicates that the primitive physical property has been ideally satisfied, and the shape *may* belong to the object category. The final result depends on the evaluation of other primitive physical properties.

It would seem that each category could simply be defined by the knowledge primitives without using the notion of functional properties. The functional property level was introduced into the representation hierarchy for two reasons. First, the subgroupings of functional properties intuitively follow the levels of named categorization typical of human concepts of function. Secondly, most functional property evaluations result in the labeling of the functional elements of the object (i.e., the portions of the structure) that fulfill the functional requirement.

Since the subcategory definition represents an increasingly specialized definition, evidence for belonging to the subcategory should result in an increased measure for the object belonging to the subcategory as opposed to just the parent category. The combination of the functional property measurement of a subcategory node, $a$, with its parent node's evaluation measure, $b$, is computed using the T-conorm:

$$S(a, b) = a + b - a \times b$$

This T-conorm is commonly referred to as the *probabilistic or* (POR) function (Bonissone & Decker, 1986). While the T-conorm is used to combine measures at a subcategory node, the final subcategory evaluation measure is actually computed as:

$$E_{subcategory} = \begin{cases} S(a,b), & \text{if } a > T, \\ 0, & \text{otherwise.} \end{cases}$$

where $T$ is a user defined threshold. Thus, the functional property measurement of a subcategory node, $a$, must be greater than some minimum in order for a shape to receive a non-zero evaluation measure for the subcategory. For the purposes of this work, a value of $T = 0$ is assumed, indicating that a shape can be assigned to a subcategory as long as there is some non-zero evidence that it meets the additional functional requirements associated with the subcategory. In practice, a final classification decision might require much stronger evidence, say $T = 0.7$, before a shape is assigned to a subcategory.

For example, to determine the overall evaluation measure of a shape for the category *straightback chair*, we first compute the overall evaluation measure for the category *conventional chair*, as previously described. The functional property *provides_back_support* is defined by 8 primitives. Denoting the measurements returned by the 8 primitives as $p_8$ through $p_{15}$, the overall evaluation measure (assuming the measure for *provides_back_support* $> T$) for the category *straightback chair* is computed as:

$$straightback\ chair ::= conventional\ chair\ \text{POR}\ provides\_back\_support$$

where

$$provides\_back\_support ::= p_8\ \text{PAND}\ p_9\ \text{PAND}\ p_{10}\ \text{PAND}\ p_{11}\ \text{PAND}\ p_{12}$$
$$\text{PAND}\ p_{13}\ \text{PAND}\ p_{14}\ \text{PAND}\ p_{15}$$

An object that can function as a *straightback chair* can also by definition function as a *conventional chair*. The T-conorm will give the object a higher evaluation measure for the subcategory *straightback chair* since there is some evidence in addition to the "minimal" amount of evidence required for the shape to belong to the parent category *conventional chair*. Thus, GRUFF performs recognition of a shape by selecting the (sub)category with the highest overall evaluation measure. This should correspond to the most specific applicable subcategory. One exception occurs when the parent category has an evaluation measure of 1 and there is non-zero evidence supporting the subcategory functional requirements. In this case, the T-conorm assigns an evaluation measure of 1 to both the category and subcategory.

The particular T-norm/T-conorm pair utilized in this paper was chosen from among representative T-norm/T-conorm possibilities (including non-probabilistic formulations) described by Bonissone and Decker (1986) after analyzing their performance in conjunction with GRUFF across a set of example shapes (Stark, Hall, & Bowyer, 1993a).

## 4. The OMLET Learning System

In this section, we describe the OMLET learning (sub)system. OMLET learns fuzzy membership functions, which are located at the leaves of an AND/OR categorization tree, from sets of training examples. OMLET works together with GRUFF to automatically learn object category definitions and use those definitions to recognize new objects.

In the training mode, OMLET uses examples to learn the fuzzy ranges for primitive measurements. Each training example consists of an object description coupled with a desired overall evaluation measure. In the testing mode, OMLET uses the previously learned ranges to act as a function-based object recognition system. Knowledge primitives form the building blocks of the OMLET system, and rules make up the representation language. The rules, which are fixed, are derived from GRUFF'S category definition tree. They indicate 1) how the knowledge primitives are combined to define functional properties, and 2) how the functional properties are combined to give the function-based definition of an object category.

Given a training example, OMLET uses the rules to construct a general proof tree for the example's given object category. The proof tree is simply a data structure that mimics the way GRUFF combines primitive evaluation measures. The proof tree also maintains the primitive ranges that are modified by the learning algorithm. An example proof tree generated from the rules that define an object in the *conventional chair* category is shown in Figure 4. The proof trees contain only those knowledge primitives which are defined using range parameters. This is because the other knowledge primitives return only 0/1 measures, and so there is no primitive membership function to learn. The training example must satisfy these "binary", or necessary, functional properties and return evaluation measures of 1 in order for the example to be a member of the given category. For example, in Figure 4, the left branch of the top PAND node represents the functional property *provides_stable_support*. This functional property is defined by a single knowledge primitive which has no range parameters. Therefore, this input to the PAND node is fixed to always return a 1.

For OMLET to obtain an overall evaluation measure for an example object, the physical measurements of the shape elements of the object are input to the primitive fuzzy membership functions in the leaves of the proof tree. The output at a leaf node represents the evaluation measure for the individual functional property. The evaluation measures are combined at the internal nodes of the tree using the probabilistic T-norm/T-conorm combiners described in Section 2.3. The overall evaluation measure of the input example is then output at the root node (see Figure 4).

Input to OMLET consists of a set of goals for specific examples from object (sub)categories. The goal includes the example's (sub)category, the elements of the 3-D shape that fulfill the functional properties, and an overall desired evaluation measure which is greater than 0 (otherwise the object is not an example of the object category). Figure 5 shows an example of a goal for a *conventional chair* object.

Using the training examples, OMLET attempts to learn the ranges used in the trapezoidal membership functions associated with the knowledge primitive definitions (see Figure 2). When a training example is presented, OMLET attempts to prove via the rule base that the object is a member of the specified category. Here, the check is to make sure the physical elements of the object listed in the goal satisfy the binary, or necessary, functional properties. So, for a *conventional chair* training example, OMLET checks that the given orientation is stable, and the given seating surface is accessible (clearance in front and above) and meets a minimum width to depth ratio. If the necessary functional properties have all been satisfied, a proof tree is constructed. The actual overall evaluation measure is then calculated in the manner described above. If the actual evaluation measure is sufficiently different from the desired evaluation measure, then the primitive fuzzy membership functions that were included in the definition need to be adjusted.

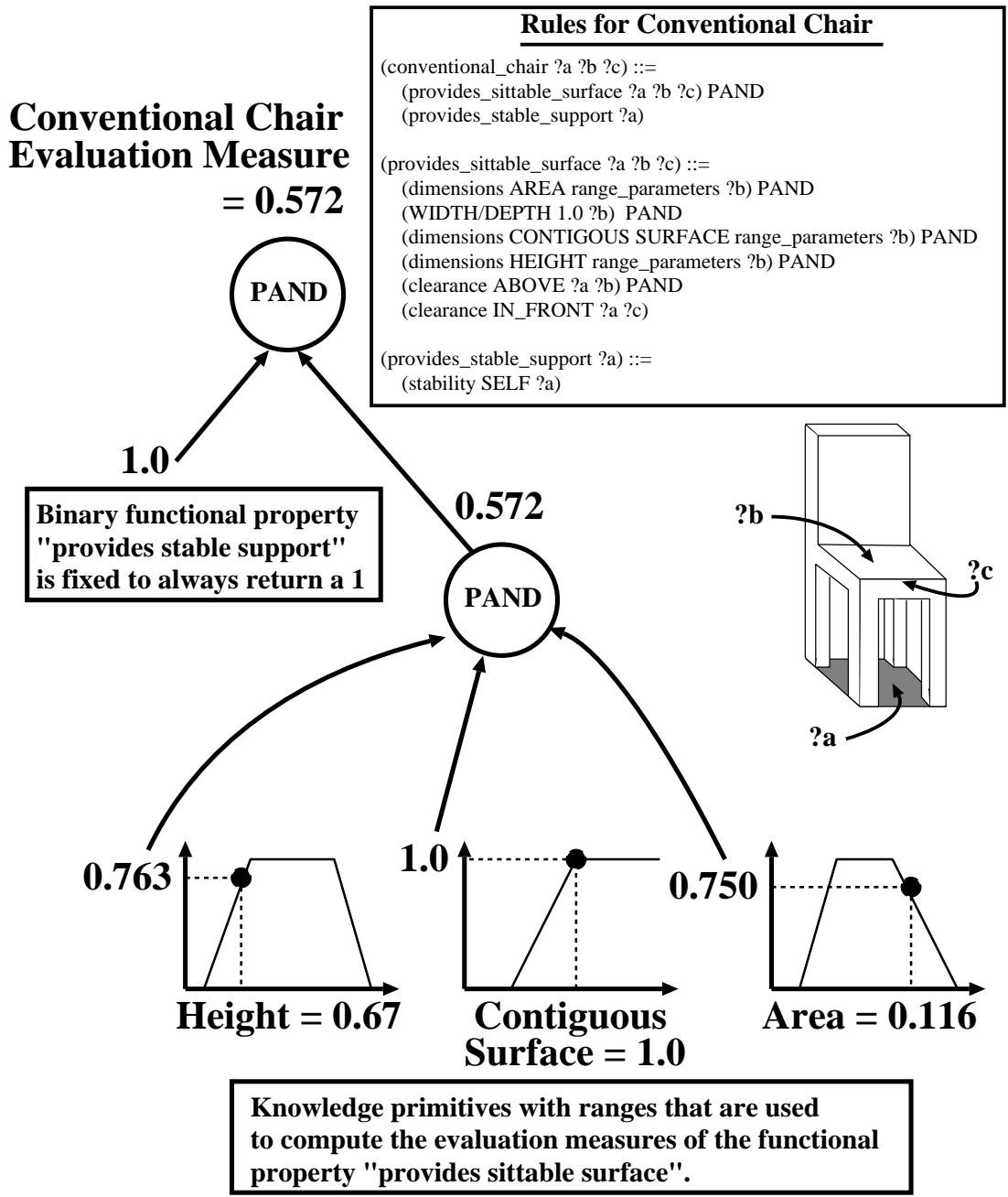

Figure 4: The simplified proof tree constructed for a learning example from the category *conventional chair*. The ?a, ?b, and ?c symbols in the rules represent the physical aspects of a shape that are used by the rules. An orientation of the shape, the face of the sittable surface, and the front edge of the sittable surface are substituted for ?a, ?b, and ?c, respectively. This way OMLET knows which elements of a shape are to be "measured" and evaluated by the knowledge primitives.

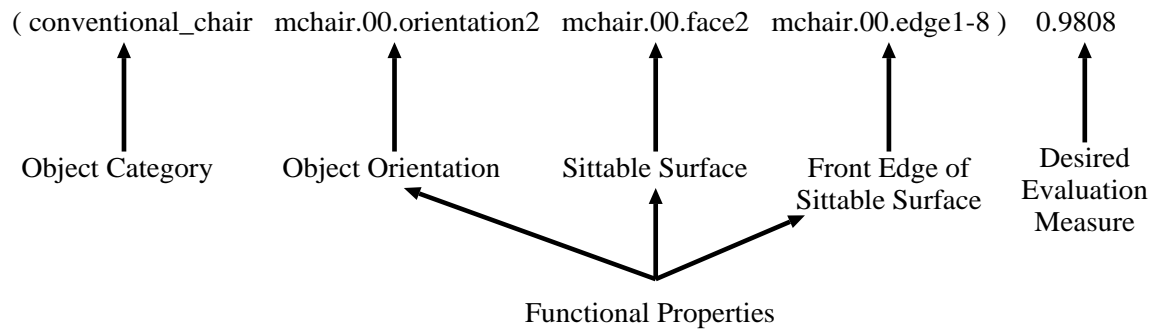

Figure 5: Training goal input to OMLET for a *conventional chair* object.

Primitive membership functions are adjusted by propagating the overall error for each training sample down through the nodes of the proof tree in a way that attempts to give each leaf node (i.e., range) some portion of the error. The range parameters ($z1$, $n1$, $n2$, and $z2$) that define the fuzzy membership trapezoids are then adjusted in an attempt to reduce the total error of the examples in the training set. The next few subsections provide details of the OMLET learning algorithm. First, we discuss the method for calculating an error value and propagating it down through the proof tree. Next, we present a method for making initial estimates of the parameters for each membership function. We describe error propagation first because it is utilized in the initialization phase. We then describe how OMLET makes adjustments to the membership functions in an attempt to reduce the error over the entire training set. The last subsection describes the general learning paradigm and provides some theoretical justification for our implementation.

### 4.1 Error Propagation

The error for a training example is defined as the difference between the desired evaluation measure and the actual evaluation measure computed by the current state of the OMLET system. A fraction of the error (defined by a "learning rate") is propagated down the proof tree through the PAND and POR nodes. Error propagation through PAND and POR nodes is handled differently. If the error at a three element PAND node is $E$, then each of the three elements will receive a portion of the error equal to the cube root of $E$ (i.e., the inverse of the PAND function). For a POR node, the full amount of error, rather than an equal share, is propagated down each link. The rationale for this treatment of error should become clear in Section 4.4.

It should be noted that while the desired evaluation measure is fed to the root of the tree and propagated down to the leaves, the error is directly computable since the actual and the projected desired values are always known at each node. The actual values at each node are those computed when the physical measurements of the object shape are fed into the leaf nodes and combined to produce an overall evaluation measure at the root. The projected desired values in the proof tree are obtained by propagating the desired evaluation measure from the root node down to the leaves. For example, given a two input PAND node with actual inputs $a_1$ and $a_2$, the actual output $A$ will be $a_1 \times a_2$ (from the T-norm in section 2.3). If the desired output of the node is $D$, then we can compute the desired inputs to the node as $d_1$ and $d_2$ by solving the following set of equations:

$$a_1 \times a_2 - d_1 \times d_2 = A - D \qquad (1)$$

and

$$a_1 - d_1 = a_2 - d_2 \qquad (2)$$

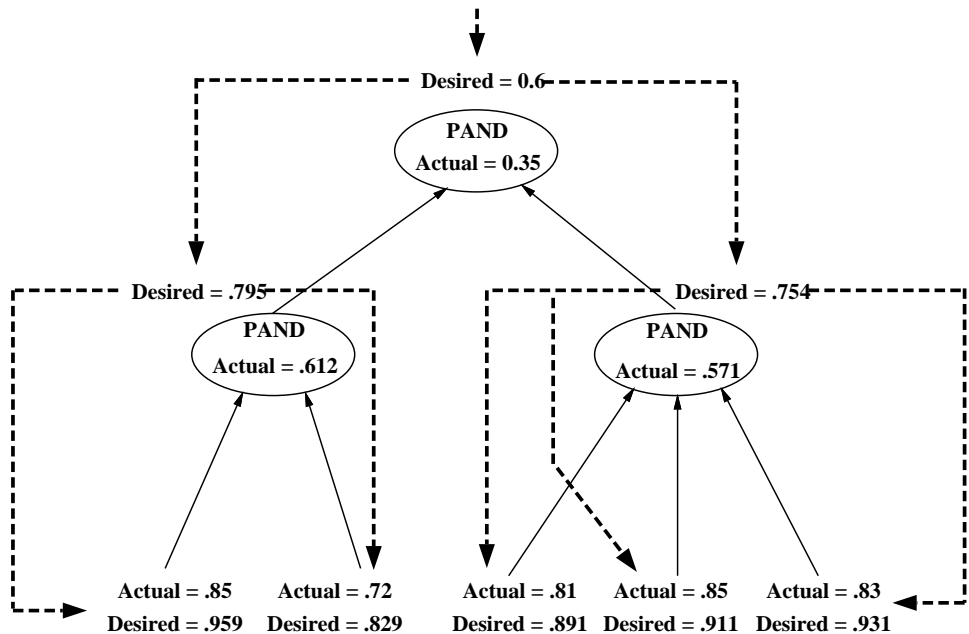

Figure 6: Example of error propagation through a PAND tree. Actual values are found when an overall evaluation measure is computed for an object. Desired values are propagated down the tree, and error is computed as $Desired - Actual$.

The first equation computes the error for the PAND node[1], while the second equation assures equal portions of the error are assigned to each input. Figure 6 shows an example of the desired values computed via Equations 1 and 2 for every node in a proof tree. In this figure, we have a known desired overall measure of $D = 0.6$ at the top PAND node, and an actual measure of $A = 0.35$ which was computed as the PAND of the actual node inputs, $a_1 = 0.612$ and $a_2 = 0.571$. Using Equations 1 and 2, we can easily compute the two unknown desired inputs $d_1$ and $d_2$ to the top PAND node (which are also the desired outputs of the bottom two PAND nodes) as 0.795 and 0.754, respectively. If there are three inputs to a PAND node, then we solve a set of three linear equations to derive the desired inputs. When there are more than three inputs to a PAND node, we divide the set of inputs recursively into groups of two or three and solve a set of two or three linear equations, respectively.

---

1. The equivalent and simpler equation $d_1 \times d_2 = D$ could be substituted here.

Since the POR nodes are used to combine a single parent category measure with a single aggregate measure for a subcategory's functional properties, there will never be more than 2 inputs to this type of node. Therefore, the full amount of error can be propagated through a POR node by simply solving the independent equations:

$$a_2 + d_1 - a_2 \times d_1 = D \tag{3}$$

and

$$a_1 + d_2 - a_1 \times d_2 = D \tag{4}$$

Eventually, some portion of the overall error is propagated to the ranges defined by the trapezoid membership functions. When the error reaches the individual ranges for a training example, the input to the primitive membership function (i.e., the x axis value) and the desired primitive evaluation measure (the y axis value) define a point that should lie somewhere on the trapezoid. We also note which leg of the trapezoid the point belongs to, based on which side of the normal portion of the range $[n1,n2]$ that the x value lies. The set of desired points for each leg can be used to make adjustments to the trapezoid in an attempt to reduce the error. OMLET collects these desired points for each leg of each membership function by propagating the error for all training examples down the proof trees. The trapezoid/range parameters ($z1,n1,n2,z2$) are adjusted at the *end* of each training epoch. Training continues for a fixed number of epochs or until some satisfactory level of performance, defined by minimal classification error rate averaged over the training set, is achieved.

### 4.2 Initial Estimate of Measurement Functions

OMLET's learning algorithm begins by making reasonable initial estimates of all fuzzy trapezoid membership functions for the physical measurements. This is accomplished by assigning actual values of 0 for the membership functions for each training example and propagating the errors (which in this case would be equal to the desired evaluation measures) down to the ranges in the leaf nodes of the proof trees. From the collections of desired points, we make an initial estimate of each trapezoidal membership function. It is only important at this stage to place the edges of the constructed normal range (the $n1$ and $n2$ range parameters) somewhere within the actual normal range. The learning algorithm will make adjustments to the $n1$ and $n2$ points on subsequent training epochs. Additionally, OMLET may set minimum or maximum limits on the values of some of the range parameters (more on this shortly).

A training example with a desired evaluation measure of 1 is considered a "perfect" example of an object from a given category. Perfect training examples are desirable in the training set because all primitive measurements for perfect examples are known to fall in the range $[n1,n2]$. For example, if a *conventional chair* training example has a desired evaluation measure of 1, then we know that all of the membership functions in its proof tree (see Figure 4) must return values of 1. This is because the result of the PAND function can be no greater than the minimum input.

Omlet now examines the set of desired points that have been propagated to each range in the definition tree and determines "limit" points. These are defined as follows. If any two desired points have $y$ values (memberships) of 1, then at least a segment of the normal range $[n1,n2]$ is known. The $n1$ range parameter is set to the minimum $x$ value of all desired points with $y$ values of 1. Similarly, the $n2$ parameter is set to the maximum $x$ value of all desired points with $y$ values of 1. Note that if only one such desired point is found then $n1$ and $n2$ are set to the same value, and the membership function is initially triangular. Since some portion of the normal range is known to be correct, an upper limit is set on the $n1$ value and a lower limit is set on the $n2$ value to assure that the known segment of the normal range is not reduced during subsequent training. Since training examples have desired membership values greater than 0, we know that all $x$ input values must lie between $z1$ and $z2$. Omlet uses the minimum and maximum $x$ values from the set of desired points to set limits on the $z1$ and $z2$ range parameters. The $z1$ range parameter is never permitted to increase above the minimum $x$ value during training. Similarly, the $z2$ value may never decrease below the maximum $x$ value in the set of desired points. Figure 7 shows the range parameters (limit points) Omlet sets during the initialization phase given a set of 10 examples.

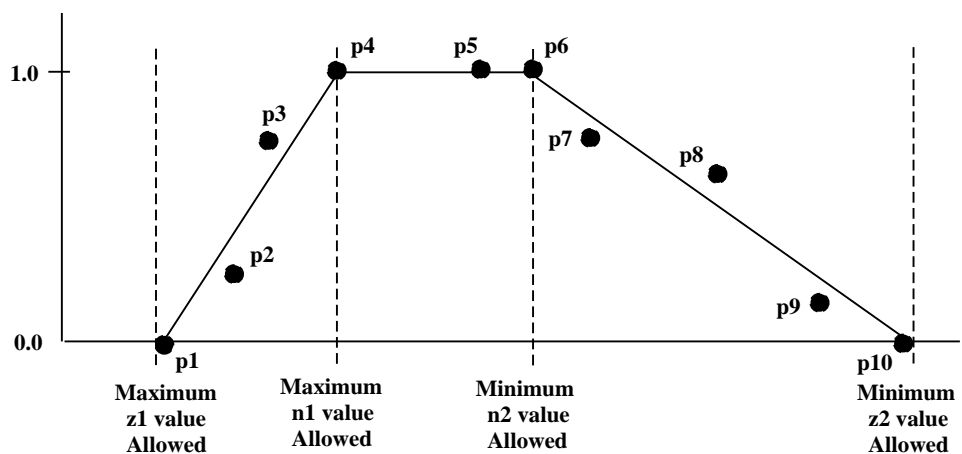

Figure 7: Range parameter limits that may be set when initializing range parameters.

The limits on the range parameters serve several purposes. First, the limits assure that perfect training examples will not be assigned evaluation measures less than 1, and that all training examples will have evaluation measures greater than 0. More importantly, by limiting the changes that can be made to some range parameters, better approximations to the desired membership functions can be learned. In subsequent learning, the error is propagated down the proof tree with the assumption that equal amounts of the error come from each input to a node. This assumption is not always valid, and there is no way to directly determine the portion of the error that belongs to each input. If an error propagated to a membership function would cause a change in one or more of the range parameters ($z1,n1,n2,z2$) that moves the parameter past its set limit, the portion of the overall error assumed to be caused by the membership function has not been correctly estimated. When this occurs the parameter is set equal to its limit, effectively reducing the degree to which changes in the membership function would compensate for the overall error. This should allow the learning algorithm to find a good solution in the case where different membership functions contribute different amounts of error.

If a segment of the normal range is known for some membership function, then initialization of the range parameters is straight-forward. The $n1$ and $n2$ values will have already been set. The $z1$ value is set simply by making the left leg of the trapezoid pass through the point $(n1,1.0)$ and the point from the set of desired points with the minimum $x$ value. Similarly, the $z2$ value is set by making the right leg of the trapezoid pass through the point $(n2,1.0)$ and the desired point with the maximum $x$ value. If there are no points to the left (right) of the $n1$ ($n2$) point, then the membership function is assumed to be one-legged (as for CONTIGUOUS SURFACE in Figure 4) and the parameters $n1$ and $z1$ ($n2$ and $z2$) are extended to a very large negative (positive) value and not permitted to change during training.

If no portion of the normal range of a membership function can be determined, then we attempt to fit a trapezoid to the set of desired points. First, the two desired points with the maximum y values are found. We assume that the normal range lies somewhere between them. A best-fit trapezoid is determined by varying the $n1$ and $n2$ range parameters over the assumed normal range, and selecting the normal range $[n1,n2]$ that produces the lowest error for the set of desired points. The error is the sum of the absolute values of the difference between the desired y value and the actual y value found for each point. The $z1$ ($z2$) range parameter is set in the same manner as before, where the left (right) trapezoid leg is forced to pass through the desired point with the minimum (maximum) x value. The $n1$ value is varied from the leftmost point of the assumed normal range to the rightmost point in small increments. For each different value of $n1$, the $n2$ value is varied from $n1$ to the rightmost point of the assumed normal range in small increments. So, we are simply testing a range of possible trapezoids (with the degree of accuracy, and number of trapezoids tested, defined by the increments in which $n1$ and $n2$ are varied) that have a normal range $[n1,n2]$ somewhere within the assumed normal range. From these we select the set of range parameters that minimize the total error over the set of training examples. The use of a best-fit trapezoid approach is helpful, because we have no initial way to accurately associate error with any given trapezoid.

### 4.3 Adjusting Membership Functions

To make adjustments to a membership trapezoid, each leg of the trapezoid is fit to a set of desired points using a least squares line fit. Recall that after every training epoch we have a set of desired points for each leg of each trapezoid. The new $z1$ ($z2$) value of the trapezoid is set to the point at which the left (right) leg intersects 0. The new $n1$ ($n2$) value is set to midway between the old $n1$ ($n2$) value and the value where the left (right) leg of the fitted line intersects $y = 1$. The new $n1$ and $n2$ values are not directly set to where the fitted trapezoid legs intersect 1 because overestimating the normal range $[n1,n2]$ can eliminate some desired points that should be used in the least squares line fit for a trapezoid leg. Desired points in the normal $[n1,n2]$ range by definition do not fall on a leg of the trapezoid, and are not used when adjusting the trapezoid legs. Therefore, if the normal range is overestimated, points that truly belong on a trapezoid leg will not be used to adjust the leg. By gradually moving the normal points $n1$ and $n2$, OMLET is better able to converge on an appropriate solution. After the new range parameter values ($z1,n1,n2,z2$) have been determined, OMLET checks to make sure that none of them lie outside any limits that may have been set in the initialization phase. Restrictions on new range parameters assure that the membership functions remain trapezoidal (or triangular if $n1 = n2$). First, $z1$ must be less than or equal to $n1$. Similarly $z2$ must be greater than or equal to $n2$. If $z1$ ($z2$) is greater (less) than $n1$ ($n2$) then $z1$ ($z2$) is set equal to $n1$ ($n2$). Also, $n1$ must be less than or equal to $n2$. In the case that there is only a single point in the set of desired points for a trapezoid leg, the leg is defined by the normal point for that leg ($n1$ for the left leg and $n2$ for the right leg) and the single desired point.

The training data may provide target points for only a portion of a trapezoid for some of the ranges. OMLET is capable of detecting this situation by observing the slope of the fitted line, and adjusting the membership function appropriately. The slope of the left trapezoid leg should be positive and the slope of the right leg should be negative. If the slope of the fitted trapezoid leg is nearly horizontal (close to 0.0), or the sign of the slope is opposite what is expected, then the normal point on that leg is moved (again, $n1$ for the left leg and $n2$ for the right leg) outward. This adjustment allows OMLET to learn one-legged membership functions, and to handle (as well as possible) situations when not enough training data is available.

A method of escaping local minima was empirically found useful. Normally OMLET does not allow a trapezoid leg to change if the change causes an increase in total error for the training set. So, it is possible for zero, one or both trapezoid legs for each range to get adjusted on an epoch. If learning slows down sufficiently, then OMLET will temporarily allow trapezoid leg changes that cause an increase in overall error in hopes of escaping a possible local minima. More precisely, if the total training set error for one epoch decreases by less than a specified threshold, then range changes that cause an increase in overall error are permitted for the next training epoch.

### 4.4 The Training Approach

In order to learn all the various subcategories defined in a category definition tree, we utilize a machine learning approach which is based on an assumption about human learning known as *one disjunct per lesson* (Lehn, 1990). Perhaps it is easiest to understand the mechanics of our learning approach if we explain the one-disjunct-per-lesson assumption in the terminology of cognitive science. Since many of the terms in machine learning are derived from the cognitive sciences, it will not be difficult to show the similarities between our algorithm and this characterization of human learning. We will also examine some of the computational characteristics of our learning algorithm that support our choice of this approach.

#### 4.4.1 One Disjunct Per Lesson

Van Lehn (1990) tells us that an effective way of teaching more complicated concepts is to build them up from simple subconcepts, as opposed to an "all-at-once" approach. For our purposes, a disjunct can be considered one of these simple subconcepts. A lesson consists of an uninterrupted sequence of demonstrations, examples, and exercises. The length of a lesson varies. Thus, we might expect a human to better understand the concept of an armchair by presenting a series of lessons, each of which introduces a single new subconcept that builds upon the previous subconcepts. For example, a first lesson teaches the concept of a conventional chair which requires only a stable sittable surface in the correct orientation. To learn what constitutes a straightback chair, we build upon the concept of conventional chair by introducing the subconcept of back support in a second lesson. So, the second lesson broadens our notion of chairs, in general. Finally, a third lesson builds upon our understanding of a straightback chair by introducing the subconcept of arm support. By contrast, the all-at-once approach may try to explain that an armchair provides a stable sittable surface in the correct orientation with some back and arm support. Here, we are trying to teach three subconcepts at one time, *and* show how the three subconcepts together form the more complex concept of an armchair. Indeed, Van Lehn (1990) cites some laboratory studies which indicate that the learning task is more difficult when more than one disjunct (subconcept) is taught per lesson.

We have chosen to utilize a machine learning algorithm which has underpinnings similar to Van Lehn's one-disjunct-per-lesson assumption. In our case, concepts and subconcepts are represented by categories and subcategories. A lesson for our algorithm consists of numerous epochs of the training examples from one (sub)category. Thus, our lesson can be viewed as an uninterrupted sequence of positive examples that "teach" the functional requirements for a single (sub)category. The length, or number of training epochs, of our lessons may vary depending on the subcategory being learned. To learn all the ranges in a category definition tree, we begin by learning the simplest concepts first. Then we learn additional more complex subconcepts by building upon the notion of the more simple concept. For example in the simplified proof tree in Figure 8, the parent category *conventional chair* will be learned before attempting to learn the subcategory (specialization) *straightback chair*. Since the subcategory *straightback chair* is itself a parent category, it will be learned before attempting to learn the even more complex subcategory *armchair*. The remainder of this subsection discusses our implementation in finer detail.

From an implementation standpoint, the simplest concepts are the functional properties associated with the categories that are directly linked to the root node in our category definition tree such as *provides_sittable_surface* and *provides_stable_support* for the category *conventional chair*. In our first lesson, we use positive examples from these "first level" (or parent) categories to learn only those membership functions associated with these categories. Once the first level categories have been learned, their membership functions are "frozen" and not permitted to change during subsequent lessons.

In our second lesson, only the membership functions of the "second level" categories (i.e., the subcategories of the first level categories in the definition tree) are learned. In Figure 8, these membership functions belong to the node *provides_back_support* for the subcategory *straightback chair*. If we have learned the "simple" functional concept associated with the parent category, the values computed for a parent category node are assumed to be reasonably accurate. For example, when the actual values in a proof tree are computed for a *straightback chair* training example, the actual values emanating from the parent category node *conventional chair* should be accurate since the concepts associated with this node have already been learned. That is, the evaluation measures for the functional properties *provides_sittable_surface* and *provides_stable_support* of a *straightback chair* example are assumed to be correct. This implies that the membership functions making up the functional requirement subtree (i.e., *provides_back_support*) are responsible for the entire error for a subcategory training example. (This explains why Equations 3 and 4 are used to propagate error through POR nodes.) Hence, the error is propagated to the modifiable leaves under a functional requirement node through a PAND subtree and learning continues as before.

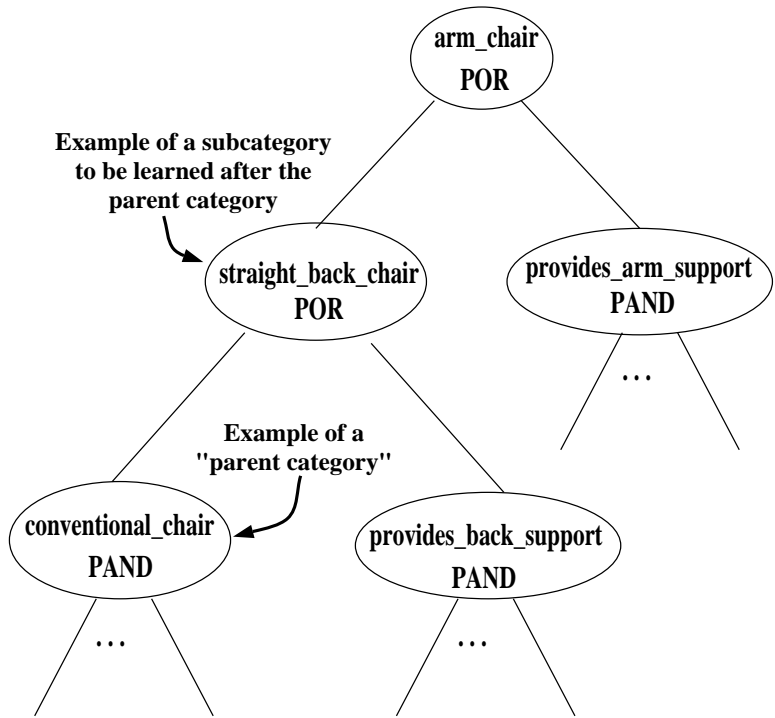

Figure 8: Simplified proof tree for an *armchair* object.

The lessons continue with each parent category being learned before any of its subcategories are learned, until all subcategories have been learned. By freezing the parent category membership functions after they have been learned, we are applying to the one-subconcept-per-lesson strategy. So in Figure 8 after learning straightback chair, the membership functions for that branch are frozen and the *armchair* subcategory is learned by modifying the membership functions under the *provides_arm_support* branch of the proof tree.

OMLET begins learning by evaluating the rule base in order to determine subcategory dependencies and assigns each (sub)category in the definition tree a level in the learning hierarchy. For example, OMLET determines that the category *conventional chair* has no parent category and its membership functions can be learned immediately (level 1). However, the evaluation measure of the subcategory *straightback chair* is dependent on the parent category *conventional chair*. The *straightback chair* subcategory is assigned to learning level 2. Subcategory *armchair* is dependent on parent category *straightback chair*, and is therefore assigned to learning level 3.

### 4.4.2 PRACTICAL JUSTIFICATION

In order to understand why we have taken a one-disjunct-per-lesson approach rather than an all-at-once approach, let's make some observations concerning how accurately blame assignment for an error can be determined for a typical training example.

Recall that error propagation through a proof tree involves projecting desired node input values from a known node output value. Consider a PAND node with a known desired output of 0.9, and two unknown inputs. We know that *both* of the inputs must be at least 0.9. This means both inputs to the PAND node fall within the relatively small range [0.9,1.0]. However, when the desired output of a two input POR node is 0.9, we can only be sure that both inputs fall in the range [0,0.9]. If the known output to the PAND or POR node is very low, say 0.1, then there is an opposite effect. That is, the unknown inputs for a POR node would lie in the relatively small range [0.0,0.1], and the unknown inputs for the PAND node would fall somewhere in the much larger range [0.1,1.0]. These observations suggest that the blame assignment for error can be propagated through a PAND node with reasonable accuracy on examples that are relatively good, say 0.7 or above. However, for high evaluation measures, an error value cannot be reliably propagated through a POR node.

Since a subcategory evaluation measure is computed as the POR of a parent category evaluation measure and the combination of additional functional requirements, all POR nodes in a proof tree have two inputs. All POR nodes (in our proof trees) have at least 1 connecting node which consists of a parent (or more general) category whose membership calculation involves only PAND connectives. The structure of the proof trees permits the membership functions which contribute to the evaluation measure of a parent category to be accurately learned prior to learning those defined in the additional functional requirements of the subcategories. That is, we can determine one of the inputs to any POR node before we attempt to propagate an error through that node. With one input *and* the desired output of a POR node known, calculation of the unknown input is trivial. Thus, our learning approach eliminates the reliability problems associated with propagating blame assignment for error through POR nodes. This will be verified in Section 6 with experimental results for the subcategories *straightback chair* and *armchair*.

The mechanics of our learning algorithm suggests that OMLET'S performance depends on how accurately blame assignment can be propagated through the PAND nodes of a proof tree. Earlier, we observed that blame assignment is less reliably propagated through PAND nodes for "bad" training examples. Not surprisingly, this suggests that the quality of the training data will have an effect on system performance. This does not mean that "bad" examples of an object (sub)category cannot, or should not, be included in the training set. Since we use a least squares line fit to adjust the fuzzy membership functions, the use of some "bad" training examples (for which the blame *may* have been inaccurately distributed among the fuzzy membership functions) should not dramatically affect the overall reliability of the learned system parameters. Rather, it is just desirable to train the system with examples that, for the most part, are good examples of their labeled object category. However, this is not unreasonable as we might expect a machine (or a human for that matter) to better learn what constitutes a chair by observing good examples of chairs.

## 5. Experimental Setup

Upon reading in the rule base, the knowledge primitive measurements of the training examples, and all training example goals, OMLET begins by learning the membership functions of all level 1 categories. The first learning epoch is used to make initial estimates of the membership functions, and then OMLET iterates for 1000 additional training epochs. A learning rate of 0.15 is used during the 1000 training epochs, so that 15 percent of the actual error for each training example is propagated to the adjustable ranges on each epoch. After the 1000 training epochs, the best range parameters (those that resulted in the lowest overall error) for level 1 categories are restored and frozen. The 1000 training epochs are then repeated for the level 2 categories, followed by the level 3 categories, and so on until all ranges in the category definition tree have been learned[2].

The performance task of the OMLET system is evaluated by how well the trained system recognizes objects that were not used in the training phase. One measurement of system performance is the error observed on the test examples. The error for a test example is computed as the absolute value of the difference between the desired and actual evaluation measures. Training/Test sets are configured two ways: random partitioning of all labeled data into training and test sets, and leave-one-out testing. In the first case, for a given size training set, 10 train/test set pairs are created by randomly partitioning all the labeled data. The error for a single test set is the average error of all test examples. The results for a given size training set are reported as the average error of the 10 partitions. In leave-one-out testing, one example in the data set is used to test while all remaining samples form the training set. This is repeated using each example in the data as the test set, and results are reported as the average error of all test examples. The average error per example versus the training set size is plotted for training sets of 10, 20, 30, ... , N-1 samples. The point with N-1 training examples represents the leave-one-out test results.

### 5.1 Test on the GRUFF Chair Database

From the evaluations of GRUFF (Stark & Bowyer, 1991), a large database of 3-D shapes specified as polyhedral boundary representations has been built up. Figure 9 shows 52 chair shapes. A number of the 52 shapes can belong to more than one category or can function in more than one stable orientation. This results in a total of 110 training examples. There are 78 labeled instances for the category *conventional chair*. Some 28 of these instances additionally satisfy the function of *straightback chair*, and 4 instances satisfy the function of *armchair*. For each shape, we have the evaluation measure for the shape's membership in different object categories, as computed by GRUFF with the hand-crafted functions for the primitive evaluation measures. This set of shapes and their evaluation measures make up the first set of training examples.

The first set of experiments will help determine how well OMLET learns a set of membership functions that minimize the overall error, and also how closely the learned membership functions approximate the original functions hand-crafted by an expert for GRUFF. A question of great practical importance to vision researchers is whether a machine learning technique can derive a set of system parameters equivalent to the hand-crafted results of the system designer. If so, the manual effort in system construction could be greatly eased. When the learning task is formulated as duplicating the GRUFF measures, the training data for these experiments is effectively "noiseless". (Noiseless in the sense that the desired evaluation measures that are used as input to OMLET are all derived in the same manner from the same set of hand-crafted fuzzy membership functions.)

---

2. In some preliminary experiments, OMLET converged on a low overall error for each level of categories anywhere between 200 and 900 training epochs. Hence the decision to train for 1000 epochs per category level. The learning rate was also determined empirically.

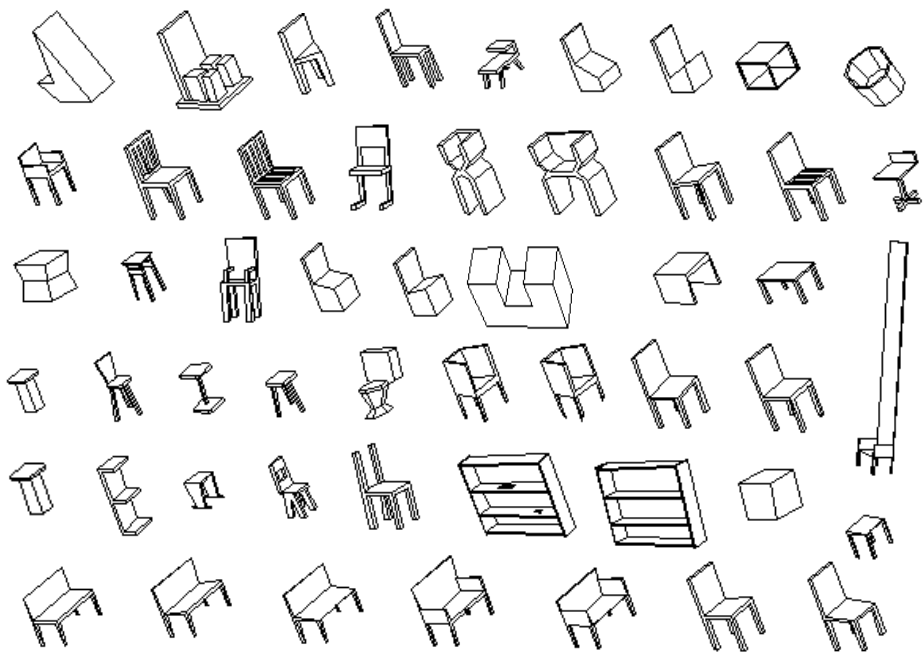

Figure 9: The 52 object chair database.

### 5.2 Test on a Synthetic Cup Database

The definition and recognition of cups is a task that has been visited frequently in machine learning research (Mitchell, Keller, & Kedar-Cabelli, 1986; Winston, Binford, Katz, & Lowry, 1983). As Winston (1983) observes, it is hard to tell vision systems what cups should look like. It is much easier to talk about the purpose and function of a cup. We convey the description of a cup by providing its functional definition. In particular, a cup is described as an object that can hold liquid, that is stable, liftable, and can be used to drink liquids. The physical identification can be made using this functional definition. In particular, for the synthetic set of objects created here, these functional properties are broken down into 19 knowledge primitives, 17 of which have range parameters.

We generated a database of 200 synthetic cup examples, for which the measurements of the knowledge primitives are randomly distributed. Hand-crafted range parameters ($z1,n1,n2,z2$) are supplied for all 17 ranges in the *cup* functional definition. To generate a cup example, a primitive measurement is randomly selected for each range. Approximately 80% of the time the primitive measurement is randomly chosen between $n1$ and $n2$. The other 20% of the time the measurement is randomly chosen outside $n1$ and $n2$, but inside $z1$ and $z2$. This cup generator program provides us with the capability to create a large number of cup examples without the time-consuming process of creating actual 3-D CAD models for each example.

### 5.3 Learning from Human Evaluation Measures

In object recognition it is important to test a system on real objects, if possible, for a number of reasons. First, we can see whether the system can approximate human judgment. Second, it is important to observe system performance in the presence of noise, which real-world data will inevitably contain. Finally, using real-world data will alleviate the need to completely hand-craft the system with synthetic data. This is actually a useful guide for the scenario where the "vision system engineer" gives the system a set of human-labeled examples, and lets the system learn the parameters. To test OMLET, we have used a set of 37 actual objects and human ratings of how well they might serve as a chair. Figure 10 shows some of the objects used in these experiments.

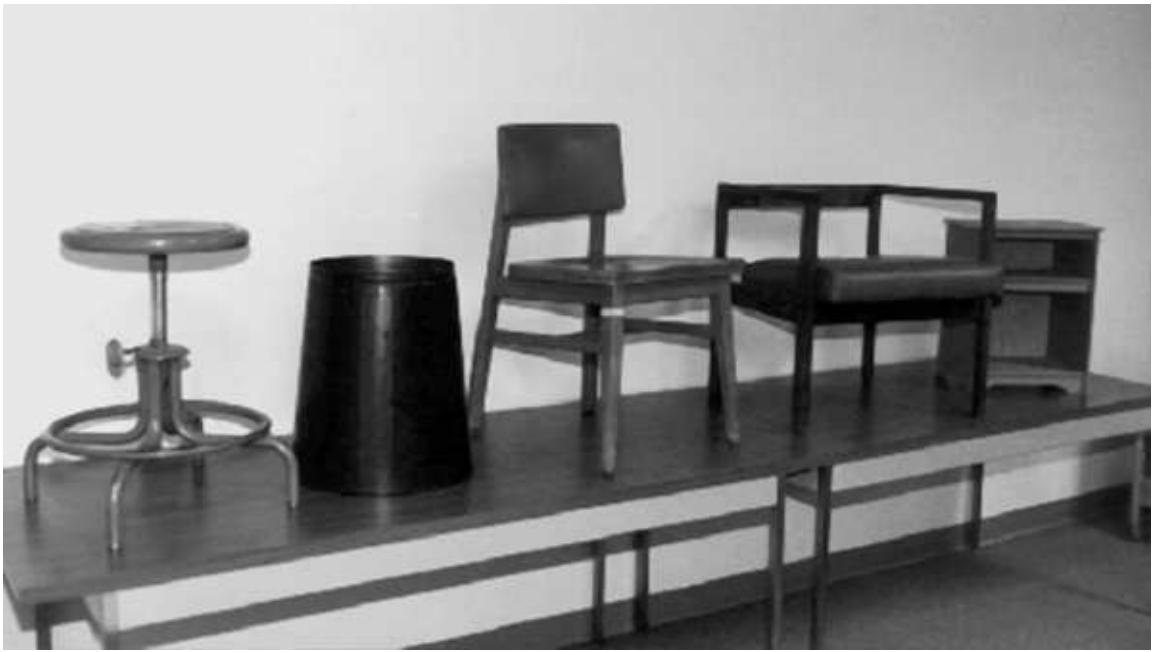

Figure 10: Some examples of the chair objects used for human evaluation tests.

In order to determine how well OMLET can learn to recognize the set of real chair-like objects, all the objects were collected together in a single room and each object was placed in the orientation in which it would most likely be recognized as a chair. For actual chairs, this is simply the orientation in which the chair would typically be used. For a metal trash can it would be an "upside down" orientation, etc. Then a group of 32 undergraduate students in an Artificial Intelligence class was given the following instructions:

> You are asked to rate each of the thirty-seven objects according to the degree of "chair-ness" that is reflected in its 3-D shape. For our purposes, "chair-ness" measures if the object could be *used as* a chair. You are to consider *only* the 3-D shape in making your rating. You should assume that each object is made of appropriate materials, so that this is not a factor in your ratings. You are to consider the suitability of the object shape only in the orientation that you see it, rather than some other orientation. Examples of factors that you should consider in rating the "chair-ness" of a shape are height, width, depth, area, relative orientation and apparent stability.
>
> You are asked to rate each shape against the requirements of three different aspects of "chair-ness". The first aspect is solely its ability to provide a stable seating surface. The second aspect is solely its ability to provide back support compatible with the seating surface. The third aspect is solely its ability to provide arm support compatible with the seat and back. Each aspect should be judged independently on a scale of 1 to 5, where 1 means it has no ability to provide the required function and 5 means that it seems ideal to provide the desired function. You may mark halfway between two numbers if you wish.

The ratings of each aspect of "chair-ness" were then averaged, normalized and rounded to the nearest multiple of 0.02 to result in values in the range [0,1]. The overall evaluation measures for the objects for the *conventional chair* category are taken as the normalized evaluation measures for the first aspect of "chair-ness", that is the object's ability to provide a stable seating surface. Overall evaluation measures for the categories *straightback chair* and *armchair* are computed using the *probabilistic or* T-conorm to combine the three aspects of "chair-ness" in the manner described in Subsection 3.3. Hence, a comfortable, sturdy chair would have a value close to 1 for "chair-ness", while the upside-down trash can has a considerably lower value (approx. 0.5).

After the objects had been rated, measurements were taken for each of the primitives describing the chair in the GRUFF system. The measurements were those required for the OMLET rules, such as the clearance from the ground, the area of the sittable surface, the height of the sittable surface, etc. Complete OMLET examples describing the objects were then created, including the aggregate evaluation measure of the objects for the categories *conventional chair*, *straightback chair*, and *armchair*. This resulted in 37 objects for the *conventional chair* category, 22 objects in the *straightback chair* category (15 objects had no back support at all), and 12 objects in the *armchair* category (10 objects that had back support did not have any arm support). There are at least two sources of noise in this experimental data: 1) the human evaluations, and 2) the actual measurements of the physical properties of the objects. For example, the standard deviations of the normalized human evaluations of the 37 objects for the *conventional chair* category are about 0.12, or 12%, on average. The results of leave-one-out testing on the 37 real-world objects are presented in the next section.

## 6. Experimental Results

There are at least four factors that may affect the performance of the Omlet system: 1) the number of training epochs, 2) the number of training samples for each category, 3) the number of ranges to be learned for each category, and 4) the quality of the training data for each category. Histograms of the desired evaluation measures of the training data are used to convey the concept of training set "quality". They are shown in Figure 11 for the Gruff chair data. The height of each histogram bin is the number of training samples with desired evaluation measures that fall within a particular range. So, the histogram of a "good" set of training data would be skewed towards the higher evaluation measures. Similarly, the histogram representing "bad" training data would be skewed towards the lower evaluation measures.

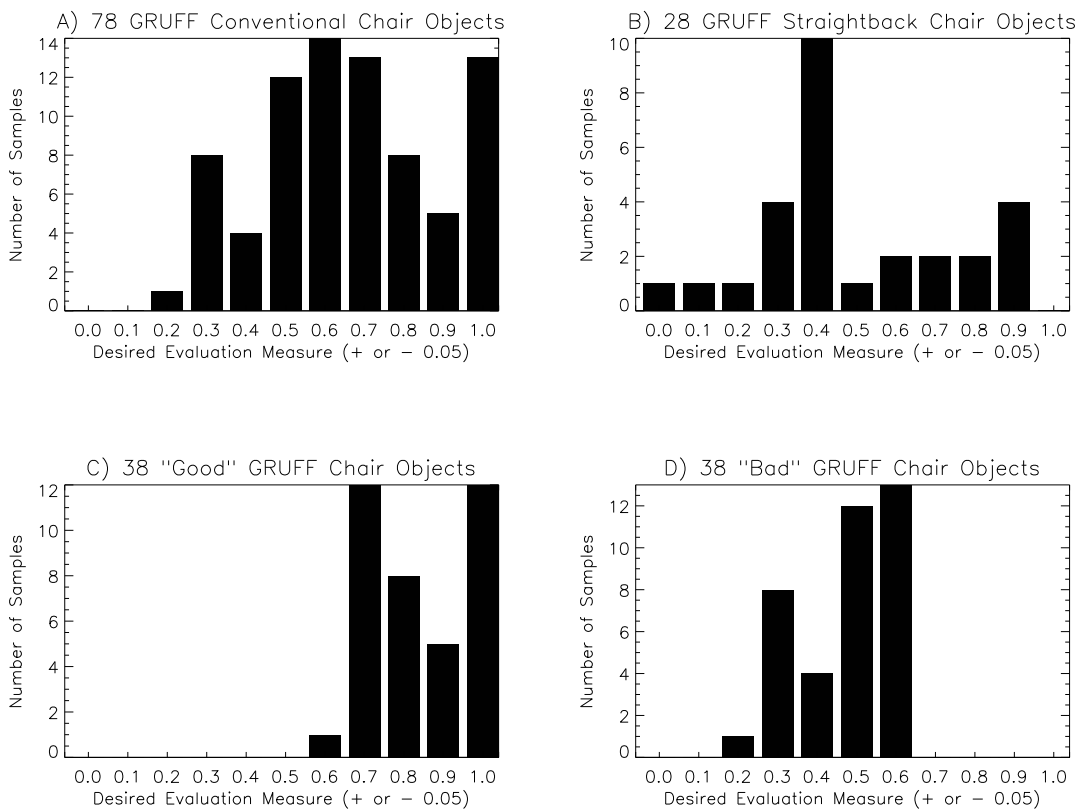

Figure 11: Histograms of desired evaluation measures of the Gruff chair training sets.

The histogram of a parent category, such as *conventional chair* or *cup*, represents the distribution of the *overall* desired evaluation measures (which are the goal measures of the examples in the data set provided as input to Omlet). However, the histograms for subcategories, such as *straightback chair* and *armchair*, represent the distributions of the desired evaluation measures associated with the additional functional requirements defined for the subcategory. For example, the histogram for the *straightback chair* category represents the quality of the *provides_back_support* portion of the *straightback chair* examples in a data set, not the overall desired evaluation measures. Recall that the ranges associated with the parent category *conventional chair* will be frozen (and presumably accurate) before learning begins for the category *straightback chair*. So, OMLET only uses *straightback chair* examples to learn the ranges associated with the *provides_back_support* functional property. Thus, when learning the ranges for the category *straightback chair*, we want to observe the quality of the back supports of the training examples. Similarly, we want to observe the quality of the arm supports of the *armchair* examples, not the overall desired evaluation measures.

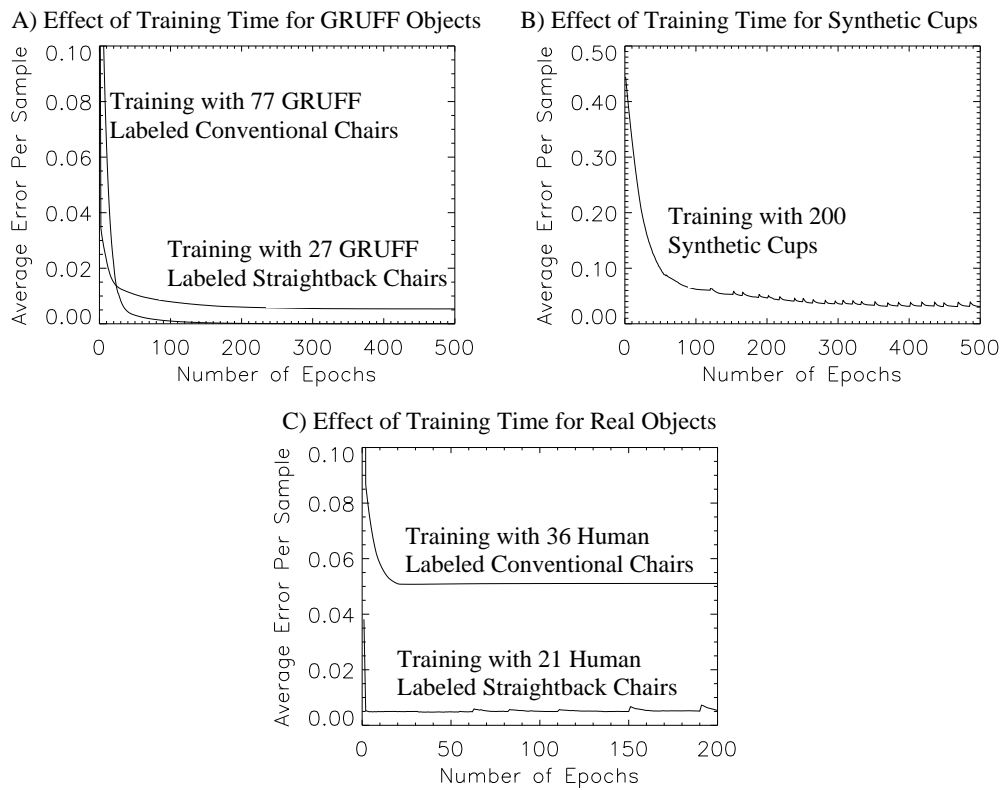

Figure 12: Average training sample error versus number of training epochs for A) GRUFF chair objects, B) synthetic cups, and C) real chair objects. These plots are for a single leave-one-out test run.

Figure 12 shows examples of the average training sample error plotted as a function of the number of training epochs for each of the three data sets (GRUFF objects, synthetic cups, and real objects). From these plots, we can see that 1000 training epochs is more than sufficient for all of the categories in the three data sets. Training could most likely be stopped after 400 epochs for any of the categories without a degradation in system performance. Since the number of training epochs is the same for all categories, and has been shown to be sufficient, we can eliminate this factor as a possible cause for the different levels of performance among categories. Some experiments in addition to those described in Section 5 were run to examine the effect of the other performance factors.

## 6.1 The Gruff Chair Database

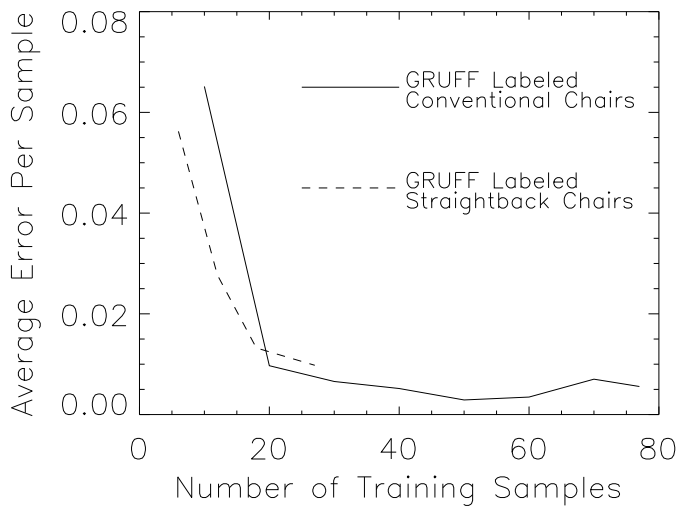

Figure 13: Omlet results for test samples from the Gruff chair database.

Figure 13 shows the plot of the average error per sample versus training set size for examples from the *conventional chair* category, and a separate plot for examples from the *straightback chair* category. Since there are only 28 *straightback chair* examples, only 3 different training set sizes (6,12,18) were evaluated in addition to the leave-one-out testing. All 78 *conventional chair* examples were used to train the ranges associated with the *conventional chair* category before the ranges for the *straightback chair* category were trained. No testing was done for the subcategory *armchair* since there were only four training samples available. The plot shows that increasing the number of training samples generally leads to a reduction in the average error. When more than 20 training examples are used, the actual evaluation measures of the test examples are within approximately 1% of the desired evaluation measures for both the *conventional chair* and *straightback chair* categories.

We should note here that the errors in overall evaluation measures found for categories at different learning levels are not directly comparable. So, the plot of the error rate for the *straightback chair* category is not directly comparable to the plot for the *conventional chair* category (Figure 13). As an example, consider an object with a desired overall evaluation measure of 0.85 for the category *conventional chair*. If Omlet computes an actual evaluation measure of 0.86, then the error for this example is 0.01. Let's assume the *provides_back_support* portion of this object has a desired evaluation measure of 0.75. The overall desired evaluation measure for this example in the category *straightback chair* would be 0.9625 (POR of 0.85 and 0.75). Now, suppose OMLET finds the actual evaluation measure for the back support of the object to be 0.76, or an error of 0.01. In this case, the actual overall evaluation measure of this example for the category *straightback chair* would be 0.9664 (POR of 0.86 and 0.76). As a result, the error of 0.01 attributed to the *provides_back_support* portion of the object is manifested as a much smaller error of 0.0039 in the overall evaluation measure of the object.

The original range parameters ($z1,n1,n2,z2$) hand-crafted by an expert for the three ranges in the conventional chair definition (see Figure 4) are:

AREA (0.057599 0.135 0.22 0.546699)

CONTIGUOUS SURFACE (0.0 1.0 1.0 1.0)

HEIGHT (0.275 0.4 0.6 1.1)

These are the range values used by GRUFF to determine the desired evaluation measures in the goals provided to OMLET. A typical example of the range parameters as learned by OMLET is:

AREA (0.057599 0.135002 0.219992 0.546706)

CONTIGUOUS_SURFACE (7.45591e-06 0.999995 10000 10000)

HEIGHT (0.275 0.400002 0.6 1.10009)

OMLET was able to determine that the CONTIGUOUS_SURFACE range was a one-legged membership function, and the $n2$ and $z2$ values (i.e., the leg that does not exist) were set to arbitrarily large values. These results show that the OMLET system is capable of using labeled examples to automatically determine range parameters which are similar to those that would be hand-crafted by an expert. This will facilitate the construction of other object category definitions.

In Figure 13, we can see that the number of training samples does indeed affect the error rate of test samples. With more than 20 or so training samples, the error rates for both the *conventional chair* and *straightback chair* categories begin to level off. So, the number of training samples becomes less of a factor affecting system performance if a sufficient number are used. What constitutes a sufficient number of training samples for a category may depend on the number of ranges to be learned and the quality of the training data. There are 3 ranges that must be learned for the category *conventional chair*, and 5 ranges that must be learned for the category *straightback chair*. The histograms of desired evaluation measures for the GRUFF *conventional chairs* and the back supports of the GRUFF *straightback chairs* in Figure 11 A and B, respectively, reflect the quality of the training data used for the leave-one-out tests.

We can isolate the effect of the quality of the training data with some additional experiments utilizing two separate data sets of GRUFF *conventional chair* examples. The number of training epochs, the number of training samples, and the number of ranges to be learned will be identical for each data set. One data set of 38 "bad" examples contains all *conventional chair* examples with desired evaluation measures less than 0.6. A second data set of "good" examples was created by selecting 38 of the remaining *conventional chair* examples. The histograms of desired evaluation measures for the examples used in the "good" and "bad" data sets are shown in Figure 11 C and D, respectively. Leave-one-out testing (37 training examples) resulted in an average error of 0.0001 for the examples in the "good" data set, and 0.1869 for the examples in the "bad" data set. Thus, it would seem that the quality of the training data has a considerable effect on the performance of the learning algorithm.

Using the set of 38 "good" *conventional chair* examples to train OMLET, the average error found using the 38 "bad" examples to test drops to 0.013 (compared to an average error of 0.1869 when 37 "bad" examples are used to train). A closer examination of the results reveals that one "bad" example contributes a relatively high error of 0.5 to the average. If this single example is excluded from the test results, the average error of the remaining 37 "bad" examples is only 0.00067. If the 38 "bad" examples are used to train OMLET, the average error found using the 38 "good" examples to test is 0.242. These results indicate that OMLET is not inherently biased to produce more accurate test results for "good" examples since we are able to achieve a low error rate for the "bad" examples when "good" training data is used. Rather, these results emphasize the importance of controlling the quality of the data used to train OMLET.

### 6.2 The Synthetic Cups Database

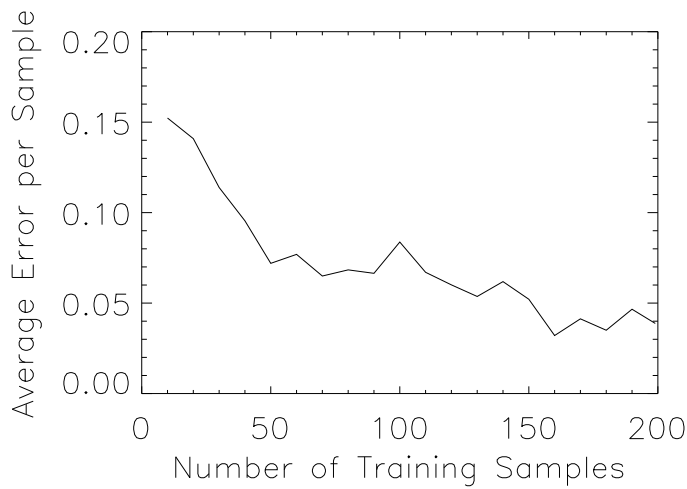

Figure 14: OMLET results for test samples from the GRUFF cup database.

Figure 14 shows the plot of the average error per sample versus training set size for examples from the randomly generated *cup* category. As before, OMLET's performance generally improves as the number of training samples is increased. A comparison of the error plots for the *conventional chair* data and the *cup* data reveals that the average error for the *cups* is higher for the same number of training samples, and the error rate decreases more erratically. The comparison of error rates between these two categories is valid since they are both at the same level in the learning hierarchy. As before, there are two performance factors that could be the cause of the different error rates. There are considerably more ranges that need to be learned for the *cup* category than for the GRUFF *conventional chair* category (17 versus 3). Also, from Figure 15 A, we can see that data set created by the cup generator program is of poor quality. Thus, due to the random nature of the synthetic cup generator program, the system was trained with shapes that, on average, are not very good examples of cups. Regardless of the poor training data, when more than 150 training samples are used, the actual evaluation measures for the *cup* test examples are within approximately 4% of the desired evaluation measures. In light of the "bad" set of shapes used as training examples and the large number of ranges that must be learned, the higher average error for cups seems reasonable.

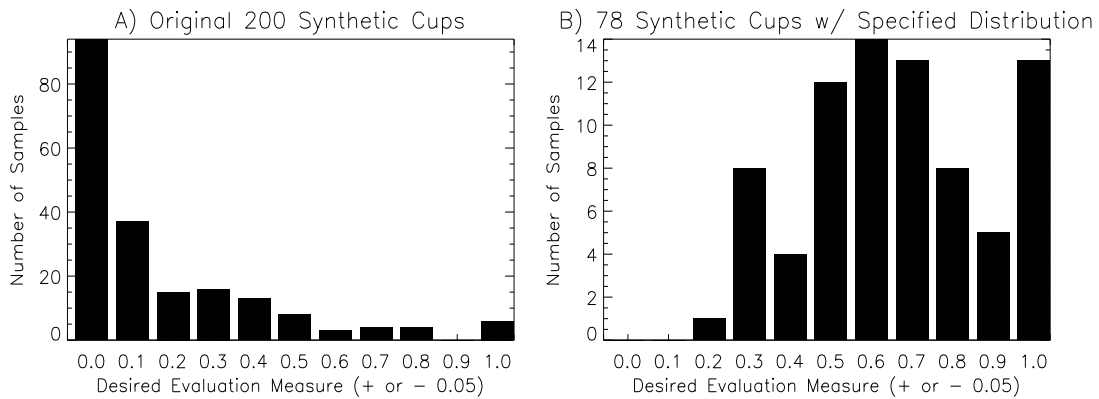

Figure 15: Histograms of desired evaluation measures of the synthetic cup training sets.

As an additional test, we generated a set of 78 synthetic cups in the same manner as before (see Section 5.2). However, we required the distribution of the desired evaluation measures of the synthetic cups to have a similar distribution as the GRUFF *conventional chair* examples (shown in Figure 11 A). Figure 15 B shows the histogram of desired evaluation measures of the examples in this second synthetic cup data set. Since the number of training epochs, the number of training examples, and the quality of the training data are the same as for the first test using the GRUFF *conventional chair* examples, this experiment isolates the effect of the number of ranges that must be learned. Performing a leave-one-out test (77 training examples), the average error per sample was found to be approximately 0.08. In Figure 13, the leave-one-out results on the 78 GRUFF *conventional chair* examples show an average error of less than 0.01 per sample. Thus, it would seem that the number of ranges to be learned affects system performance considerably.

| (Sub)Category | Number of Training Samples | Average Desired Evaluation Measure | Average Error per Sample |
|---|---|---|---|
| Conventional Chair | 36 | 0.8447 | 0.0715373 |
| Straightback Chair | 21 | 0.9927 | 0.0066456 |
| Armchair | 11 | 0.9973 | 0.0022430 |

Table 1: Leave-one-out test results for real-object database with evaluation measures derived from human ratings of the objects.

Finally, we created a set of 200 synthetic cups with a similar distribution as the GRUFF *conventional chair* examples. The histogram of desired evaluation measures of the examples in this third synthetic cup data set would look similar to the histograms in Figure 11 A, and Figure 15 B. Performing a leave-one-out test (199 training examples), the average error per sample was found to be approximately 0.023. Compared to the error rate of the original 200 synthetic cups (approximately 0.04), we again note that "better" training data improved system performance considerably. Compared to the error rate of the 78 synthetic cup data set (approximately 0.08), which is similar in quality, we see the increased number of training samples significantly improved system performance. The error rate for this third synthetic cup data set with 200 examples is still higher than the error rate for the GRUFF data set of 78 *conventional chair* objects (less than 0.01), which has a similar quality distribution. Consider that for the GRUFF data set we used 77 training examples to learn the 3 ranges of the *conventional chair* category, and for the synthetic cup data set, we used 199 training examples to learn the 17 ranges of the *cup* category.

### 6.3 The Chair Database for Human Evaluation

Leave-one-out test results for the real-object database with evaluation measures derived from human ratings of the objects are listed in Table 1. Recall that the error rates are not directly comparable among the three categories. The actual evaluation measures for the *conventional chairs* objects are within approximately 7% of the human evaluation measures. The average error here is about 6% greater average error than for the GRUFF data with a similar number of training samples. The histogram in Figure 16 A shows that the data set of real *conventional chair* objects contains mostly "good" examples. Thus, the higher average error can probably be attributed to the "noise" associated with the real-object evaluation measures. Considering an average standard deviation of 12% for the human evaluations of the *conventional chair* objects, a 7% average error per sample for the OMLET results does not seem unreasonable. The actual evaluation measures for the real-object *straightback chairs* and *armchairs* differ on average by less than 1% from the desired measures. As before, all *conventional chair* examples were used to train the ranges associated with the *conventional* *chair* category before the ranges for the *straightback chair* category were trained. The histograms of desired evaluation measures for the back support of the real *straightback chair* objects and the arm support of the real *armchair* objects are shown in Figure 16 B and C, respectively.

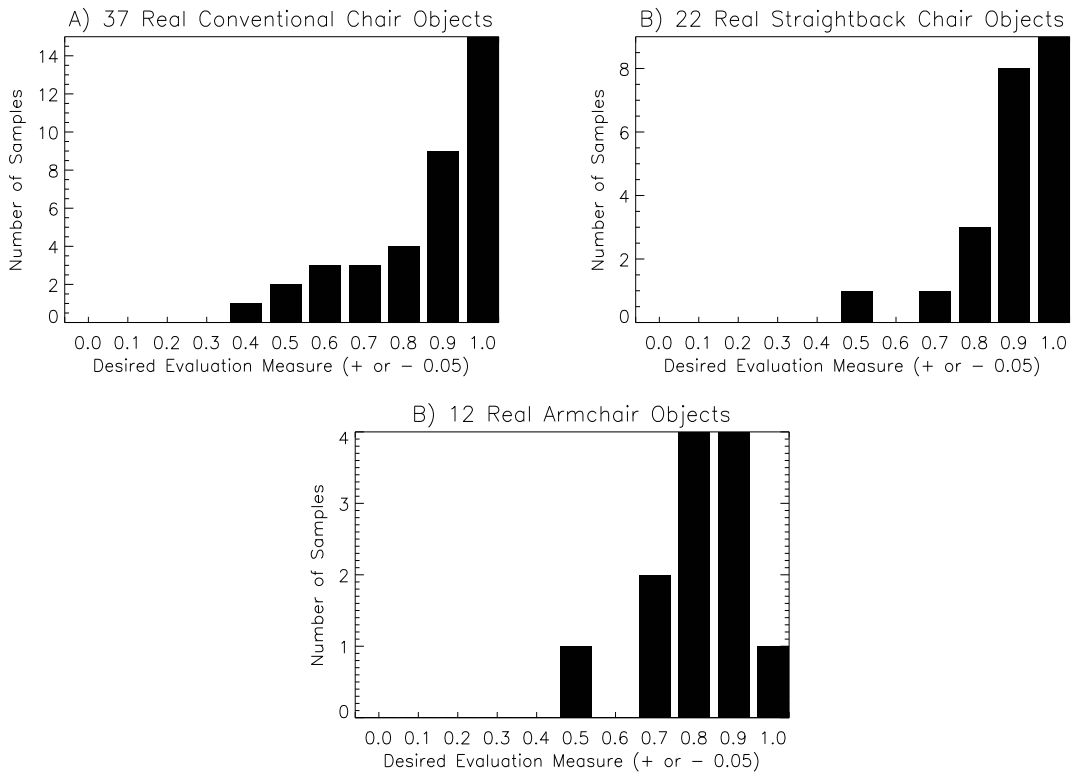

Figure 16: Histograms of desired evaluation measures of the real-object training sets.

## 7. Summary and Discussion

We have presented a system (OMLET) which uses labeled training examples to learn fuzzy membership functions embedded in a function-based object recognition system. The fuzzy membership functions are used to provide evaluation measures which determine how well a shape fits the functional description of an object category. The OMLET system is an example of using machine learning techniques to aid in the development of a computer vision system. We have shown that it is possible to accurately and automatically learn system parameters which would otherwise have to be provided by a human expert. OMLET may be used to aid in the construction of other object categories for the GRUFF object recognition system. The expert does not need to concentrate on "hand-tweaking" the range parameters to improve system performance, but rather on providing a good set of example objects to "show" to OMLET. This is intuitively appealing in that we are deriving descriptions of objects we would like GRUFF to recognize by providing examples from the object category. Additionally, we have been able to demonstrate that the performance of the learning algorithm is affected by the number and quality of the training examples.

It should be possible for the learning approach described in this paper to be applied to other systems in which measurements (or other values) are combined in a tree structure. All cases are covered by our approach, except the case of 2 leaves leading directly to a POR node. However, a generalization of our method for treating POR nodes may be developed to handle this situation. The tree structure in our CV system is composed entirely of *probabilistic and* and *probabilistic or* nodes, which are used to combine measurements. It is possible that a similar approach is applicable to tree structures in which other types of nodes (T-norms or T-conorms) are used.

The OMLET system should make it easier to adapt the GRUFF system to new object domains. Early versions of GRUFF performed object recognition starting from complete 3-D shape descriptions (Stark & Bowyer, 1991, 1994; Sutton et al., 1993) rather than from real sensory data. The task of reliably extracting accurate object shape descriptions from normal intensity images is beyond the current state of the art in computer vision. Although work in, for example, binocular stereo, is steadily progressing, accurate models of object shape are more readily extracted from range imagery. Whereas in normal imagery a pixel value represents the intensity of reflected light, in range imagery a pixel value represents the distance to a point in the scene. A version of GRUFF has been developed which attempts to recognize object functionality from the shape model that is extracted from a single range image (Stark, Hoover, Goldgof, & Bowyer, 1993b). A major difficulty here is, of course, that a single range image does not yield a complete model of the 3-D shape of an object. The "back half" of the object shape is unseen (Hoover, Goldgof, & Bowyer, 1995). The accumulation of a complete 3-D shape model through a sequence of range images is a topic of current research. If this problem was solved, then it is conceivable that an OMLET training example might consist of a sequence of range images along with some operator annotations to identify which portions of the images correspond to the functionally important parts of the object (seating surface, back support surface, etc.).

## Acknowledgements

This research was supported by Air Force Office of Scientific Research grant F49620-92-J-0223 and National Science Foundation grant IRI-91-20895.